\documentclass[runningheads]{llncs}
\usepackage{graphicx}
\usepackage{romannum}
\usepackage{caption}
\usepackage{multirow}
\newcommand{\repeatthanks}{\textsuperscript{\thefootnote}}
\usepackage{hyperref, xcolor}
\begin{document}
\pagenumbering{arabic}
\title{Graph data modelling for outcome prediction in oropharyngeal cancer patients}
\titlerunning{Graph data modelling for outcome prediction in OPC patients}
\author{Nithya Bhasker \inst{1} \and
Stefan Leger \thanks{work carried out during their time at \inst{1}} \inst{1} \and
Alexander Zwanenburg\inst{1,2,3} \and
Chethan Babu Reddy \repeatthanks \inst{1} \and
Sebastian Bodenstedt\inst{1} \and
Steffen Löck \thanks{shared last authorship}\inst{3} \and
Stefanie Speidel \repeatthanks\inst{1}}
\authorrunning{Nithya Bhasker et al.}
\institute{Department of Translational Surgical Oncology, National Center for Tumor Diseases (NCT/UCC) Dresden, Fetscherstraße 74, 01307 Dresden, Germany: German Cancer Research Center (DKFZ), Heidelberg, Germany; Faculty of Medicine and University Hospital Carl Gustav Carus, Technische Universität Dresden, Dresden, Germany; Helmholtz-Zentrum Dresden-Rossendorf (HZDR), Dresden, Germany \and
National Center for Tumor Diseases (NCT/UCC) Dresden, Dresden, Germany \and
OncoRay - National Center for Radiation Research in Oncology, Faculty of Medicine and University Hospital Carl Gustav Carus, Technische Universität Dresden, Helmholtz-Zentrum Dresden-Rossendorf, Dresden, Germany \\
\email{nithya.bhasker@nct-dresden.de}}
\maketitle              % typeset the header of the contribution
\begin{abstract}
Graph neural networks (GNNs) are becoming increasingly popular in the medical domain for the tasks of disease classification and outcome prediction. Since patient data is not readily available as a graph, most existing methods either manually define a patient graph, or learn a latent graph based on pairwise similarities between the patients. There are also hypergraph neural network (HGNN)-based methods that were introduced recently to exploit potential higher order associations between the patients by representing them as a hypergraph. In this work, we propose a patient hypergraph network (PHGN), which has been investigated in an inductive learning setup for binary outcome prediction in oropharyngeal cancer (OPC) patients using computed tomography (CT)-based radiomic features for the first time. Additionally, the proposed model was extended to perform time-to-event analyses, and compared with GNN and baseline linear models. 

\end{abstract}
\section{Introduction}
\setcounter{footnote}{0}
Radiomics holds great potential for outcome prediction and stratification of patients into different risk groups in order to personalise cancer therapy \cite{aerts2014decoding,lambin2017radiomics}. The use of traditional machine learning algorithms for radiomic risk modelling, like Cox or logistic regression, is well established in literature \cite{leger2017comparative,vallieres2017radiomics}. Recent advances include convolutional neural network (CNN) - based methods \cite{lombardo2021distant,starke20202d}, which use entire slices or volumes. However, these methods might require substantial training data to significantly outperform the traditional methods \cite{lin2020comparison}.

Graph neural networks (GNNs) using population graph based strategies \cite{cosmo2020latent,kazi2019graph,parisot2018disease} are a promising approach for disease classification. These methods define a population graph from available patient data (e.g.: radiological imaging and clinical features) to exploit structural information between patients, and use GNNs for classification. Early population graph based methods \cite{cosmo2020latent,parisot2018disease} connected patients using forms of pairwise similarities between them. As a result, each edge in such a population graph connected only two patients (nodes) and could not capture potential higher order associations among the entire cohort. In order to overcome this limitation, hypergraph neural network (HGNN) based methods, which connected two or more patients by hyperedges, were proposed. 

Most existing population HGNN methods make use of magnetic resonance imaging (MRI) \cite{madine2020diagnosing} or histopathological imaging features \cite{di2022big}. Since computed tomography (CT) imaging is a part of the standard treatment planning procedure for cancer patients undergoing radiotherapy, it would be beneficial to predict risk for the patients at this stage by making use of their CT features. Therefore in this work, we build a novel HGNN model that uses CT imaging features for outcome prediction and evaluate it using oropharyngeal cancer (OPC) datasets exemplarily.

The main contributions of this study are threefold - (\romannum{1}) to develop a patient hypergraph network (PHGN) using CT based radiomic features, (\romannum{2}) to extend this model to perform time-to-event analyses, and (\romannum{3}) to compare the performance of the developed model with that of a GNN and baseline linear model for the tasks of binary classification and time-to-event analyses. All the models are trained in an inductive learning setup. The purpose of this study is to systematically investigate the potential of representing the patient cohort and their associated radiomic features as a graph or a hypergraph for outcome prediction.

%TODO:
% - check di2021hypergraph
\section{Patient cohorts and feature pre-processing}
The data used for this study was sourced from three independent datasets available on the cancer imaging archive (TCIA) \cite{clark2013cancer} - (\romannum{1}) OPC-Radiomics (OPC) \cite{kwan2018radiomic,opc}, (\romannum{2}) Head-Neck-PET-CT (HNPC) \cite{vallieres2017radiomics,hnpc}, and (\romannum{3}) Head-Neck-Radiomics-HN1 (HN1) \cite{aerts2014decoding,hn1}. The pre-treatment CT and delineated primary gross tumour volume (GTV) from these datasets were used to extract statistical, morphological, and texture-based features in compliance with the image biomarker standardisation initiative (IBSI) \cite{zwanenburg2019assessing} using the medical image radiomics processor framework (\url{https://github.com/oncoray/mirp}).

The OPC and HNPC datasets were merged and used as exploratory cohort in a five-fold cross validation setup, while the HN1 dataset was used as an independent test cohort. The endpoints for time-to-event analyses were overall survival (OS) and freedom from distant metastasis (DM). For classification, the OS data was binarised using a specific threshold \cite{parmar2015radiomic}, i.e., a positive label was assigned to the patients who did not survive beyond the threshold duration, which in this study was set to two years as it was close to the median survival duration of the exploratory cohort with an event (786 days). The other endpoint for binary classification was human papillomavirus (HPV)/p16 status because of its significant impact on the outcome of OPC patients \cite{spence2016hpv}, and a non-invasive method of determining the HPV status could help in assigning risk to OPC patients with unknown HPV status \cite{lang2021deep}. 
 
The selection criteria for the exploratory and test cohort were - (\romannum{1}) well defined CT and corresponding primary GTV segmentation, (\romannum{2}) successful feature extraction, (\romannum{3}) no multiple GTVs, and (\romannum{4}) carcinoma at oropharynx subsite with known HPV status. For the binary OS classification, patients who had not experienced an event and were lost to follow up before the set threshold were excluded. For time-to-event analyses, patients who experienced an event at an unknown time point were excluded. An overview of endpoints for different cohorts are summarised in Table \ref{pat_char}.

\begin{table}[t]
\centering
\caption{An overview of endpoints for different cohorts. For the HPV task, the event is a positive HPV status (label 1). Time indicated in the table is the median time period between the date of diagnosis and the occurrence of event for patients who experienced an event. Label 1 indicates the occurrence of an event. Abbreviations: Expl.: Exploratory}\label{pat_char}
\begin{tabular}{|c|c|c|c|c|c|c|c|c|}
\hline
& \multicolumn{2}{c}{HPV} & \multicolumn{2}{|c}{bin OS} & \multicolumn{2}{|c}{OS} & \multicolumn{2}{|c|}{DM} \\
\cline{2-9}
& Expl. & Test & Expl. & Test & Expl. & Test & Expl. & Test \\
\hline
size, $n$ & 519 & 80 & 510 & 80 & 519 & 80 & 518 & 80 \\
\hline
Event (0/1) & 144/375 & 57/23 & 414/96 & 61/19 & 319/200 & 34/46 & 449/69 & 73/7 \\
\hline
Median Time (days) & - & - & - & - & 786 & 1041.5 & 440 & 532 \\
\hline
\end{tabular}
\end{table}

The extracted features of the exploratory cohort were standardised to have zero mean and unit variance. The mean and scale calculated for the exploratory cohort were used to standardise the features of the test cohort. Thereafter, highly correlated features in the exploratory cohort (Spearman correlation co-efficient, $\rho >0.9$) were grouped into singleton and non-singleton clusters using a hierarchical feature clustering technique \cite{leger2017comparative}. Each cluster was represented by a feature that was most correlated to all other features in that cluster \cite{tolocsi2011classification}. The test cohort was subsequently transformed to contain the same set of features as the exploratory cohort. 

Furthermore, feature selection was performed to retain only the most relevant features. The exploratory cohort was resampled with replacement to generate $100$ bootstrap samples. Normalised feature scores were calculated for each bootstrap sample using - (\romannum{1}) mutual information for binary classification \cite{kozachenko1987sample}, and (\romannum{2}) co-efficients of Cox proportional hazard model with elastic-net penalty \cite{simon2011regularization} for time-to-event analyses. The features in the exploratory and test cohort were later ranked in the order of the obtained cumulative absolute feature scores. Features with non-zero scores were considered relevant and retained.

%TODO:
% justify selection criteria
% what do the co-efficients of Cox model mean?
%- why HPV endpoint?

%whether contrast enhanced?
%little more about GTV
%how was HPV assessed - p16 method or ...?
%should you normalise the feature score before ranking

\section{Classification and risk models}
The performance of two GNN models and one baseline linear model were assessed for each task. The first GNN model used in the study was an interpretation of the latent patient network learning (LPNL) proposed by Cosmo et al. \cite{cosmo2020latent}, and the second GNN model was a custom patient hypergraph network (PHGN) built using hypergraph convolutional layers \cite{bai2021hypergraph}. 

\subsection{Construction of patient graphs}
The input graph for the LPNL model was learnt during training in an end-to-end manner. Each patient denoted a node in this graph with their features contained as node features. The features of the patients were projected onto a latent space using a multilayer perceptron (MLP) network. The edges between the nodes were computed by using a soft threshold on the pairwise distances between the patient features in the latent space as described in the work of Cosmo et al. \cite{cosmo2020latent}.  

In order to capture potential higher order associations between the patients, a hypergraph was constructed for the PHGN model. Each node in the hypergraph was represented by an individual patient. The features of the patients were projected onto a latent space using a fully connected layer, and the $k$ nearest neighbours were identified for each node based on the latent representation of the features. The parameter $k$ was chosen using hyperparameter tuning on the exploratory cohort. Hyperedges were subsequently established between a node and its $k$ nearest neighbours. The latent representation of features was considered as node features. 

\begin{figure}[t]
    \centering
    \includegraphics[width=\textwidth]{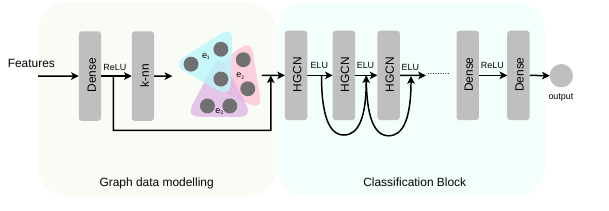}
    \caption{Architecture of the PHGN model. Abbreviations - ReLU: rectified linear unit, k-nn: k-nearest neighbours, $e_{n}$ - hyperedges, HGCN: hypergraph convolutional network, ELU: exponential linear unit.}
    \label{models}
\end{figure}

The graphs were then fed to the classification block which consisted of a series of graph/ hypergraph convolutional layers, followed by fully connected layers. To avoid overfitting in the case of PHGN model, the hidden node features were normalised and residual connections were established between the hidden layers \cite{zhou2021understanding}. Figure \ref{models} illustrates the architecture of the PHGN model. More implementation details for both GNN models can be found in the supplementary material.

%TODO
% Figure 
%add some equations if necessary
%motivation for cosine distance, knn

\subsection{Binary classification}
The baseline model used for binary classification was a logistic regression classifier with elastic-net penalty. The elastic-net penalty is a combination of both l1 and l2-norm penalties, and the ratio between these penalties is chosen using hyperparameter tuning. The input to the model were features of the patients, and the output were probability estimates for the positive class. For the GNN models, a sigmoid function was applied to the output to obtain the probability estimates for the positive class. 

%TODO:
%logistic regression citation

\subsection{Time-to-event analyses}
A Cox proportional hazard model with elastic-net penalty was used as the baseline model for time-to-event analyses  \cite{simon2011regularization}. The model accepted patient features and produced a risk estimate of experiencing an event for each patient. The GNN models were extended to perform time-to-event analyses using DeepSurv method \cite{katzman2018deepsurv}, which is a deep learning based Cox proportional hazard model. The parameters of the Cox model were the weights of the GNNs. Given the features of a patient, the models learnt to estimate the risk that the patient experiences an event. This was achieved by minimising an average negative log partial likelihood loss function with regularisation \cite{katzman2018deepsurv}. 

%TODO:
%negative log likelihod
%what is the risk estimate
%is bias required to be zero in the output layer?

\section{Experiment design}
All the models were trained in a supervised manner using a subset of ranked patient features. The size of the subset was determined by hyperparameter optimisation. The hyperparameters of all the models were tuned on a five-fold cross validation setting of the exploratory cohort. The best five models were chosen based on the average training and validation scores. All five models were used to obtain predictions on the test set, and the mean test predictions were used to calculate the performance metrics (reported in section \ref{results}). For binary classification, a threshold of $0.5$ was applied on the predictions to obtain the class labels. Furthermore, the predictions from the baseline, LPNL and PHGN classification models were averaged to obtain ensemble model predictions (Combo) and calculate the performance metrics thereof. 

The performance metrics used for classification were sensitivity, specificity, area under receiver operator characteristic curve (AUC), F1 score and accuracy. The models performing time-to-event analyses were evaluated using Harrell's concordance index (c-index). To address the class imbalance problem in binary classification, the minority class of the training set of each fold was oversampled using an adaptive synthetic oversampling technique (ADASYN) \cite{Adasyn}. In addition to addressing the class imbalance problem, the ADASYN technique reduces bias by generating samples that are harder to learn. 

The GNN models used Adam optimiser during training. The learning rate and weight decay parameters for the optimiser were chosen based on hyperparameter tuning. The loss function used for classification was binary cross entropy loss, and for time-to-event analyses a custom loss as implemented in DeepSurv \cite{katzman2018deepsurv} was used. The LPNL model was trained in batches of size $128$ for binary classification to limit the load on GPU memory. 

%TODO:
%-explain a little bit more about the concordance index. 

Kaplan-Meier (KM) survival analyses were performed for different risk groups of the test cohort. The risk groups were formed based on the predictions of the models. A threshold of $0.5$ was used for binary survival outcome to stratify the patients. In the case of time-to-event analyses, the threshold used to stratify patients was calculated based on validation risk predicted by the models for each fold \cite{lombardo2021distant}. A log-rank test was performed to compare the two risk groups. The difference between the risk groups with a p-value of $p<0.05$ was considered statistically significant.

\section{Results}
~\label{results}
The test results for the binary classification are summarised in Table \ref{bin_tab}. For HPV status prediction, the baseline linear model showed the best performance, while PHGN model showed the worst performance. The combination of models could improve the AUC by a slight margin. All models showed similar performance for binary OS prediction. In comparison to other models, the PHGN model could achieve a higher specificity score, which resulted in higher F1 and accuracy scores. On performing KM survival analysis, all models were able to achieve significant stratification based on binary OS (refer supplementary material). 

\begin{table}[t]
\caption{Binary classification results. Performance metrics for the test cohort are calculated using the average test predictions (probability estimates for the positive class) over the five cross validation folds. Abbreviations - Sens.: Sensitivity, Spec.: Specificity, Acc.: Accuracy.\\}\label{bin_tab}
\begin{minipage}{0.45\textwidth}
\centering
\begin{tabular}{|l|l|l|l|l|}
\hline
Metric &  Linear & LPNL & PHGN & Combo\\
\hline
AUC & 0.76 & 0.74 & 0.64 & \textbf{0.78}\\
Sens. & \textbf{0.83} & \textbf{0.83} & \textbf{0.83} & \textbf{0.83}\\
Spec. & \textbf{0.60} & 0.49 & 0.37 & 0.56\\
F1 & \textbf{0.65} & 0.58 & 0.50 & 0.63\\
Acc. & \textbf{0.66} & 0.59 & 0.50 & 0.64\\
\hline
\end{tabular}\\
(a) HPV status prediction
\end{minipage}
\hfill
\begin{minipage}{0.45\textwidth}
\centering
\begin{tabular}{|l|l|l|l|l|}
\hline
Metric &  Linear & LPNL & PHGN & Combo\\
\hline
AUC & \textbf{0.75} & 0.72 & 0.72 & \textbf{0.75}\\
Sens. & \textbf{0.84} & \textbf{0.84} & 0.63 & 0.79\\
Spec. & 0.54 & 0.52 & \textbf{0.72} & 0.61\\
F1 & 0.59 & 0.58 & \textbf{0.64} & 0.62\\
Acc. & 0.61 & 0.60 & \textbf{0.70} & 0.65\\
\hline
\end{tabular}\\
(b) Binary OS prediction 
\end{minipage}
\end{table}

\begin{figure}[t]
    \begin{minipage}{\textwidth}
    \centering
    \includegraphics[width=0.32\textwidth]{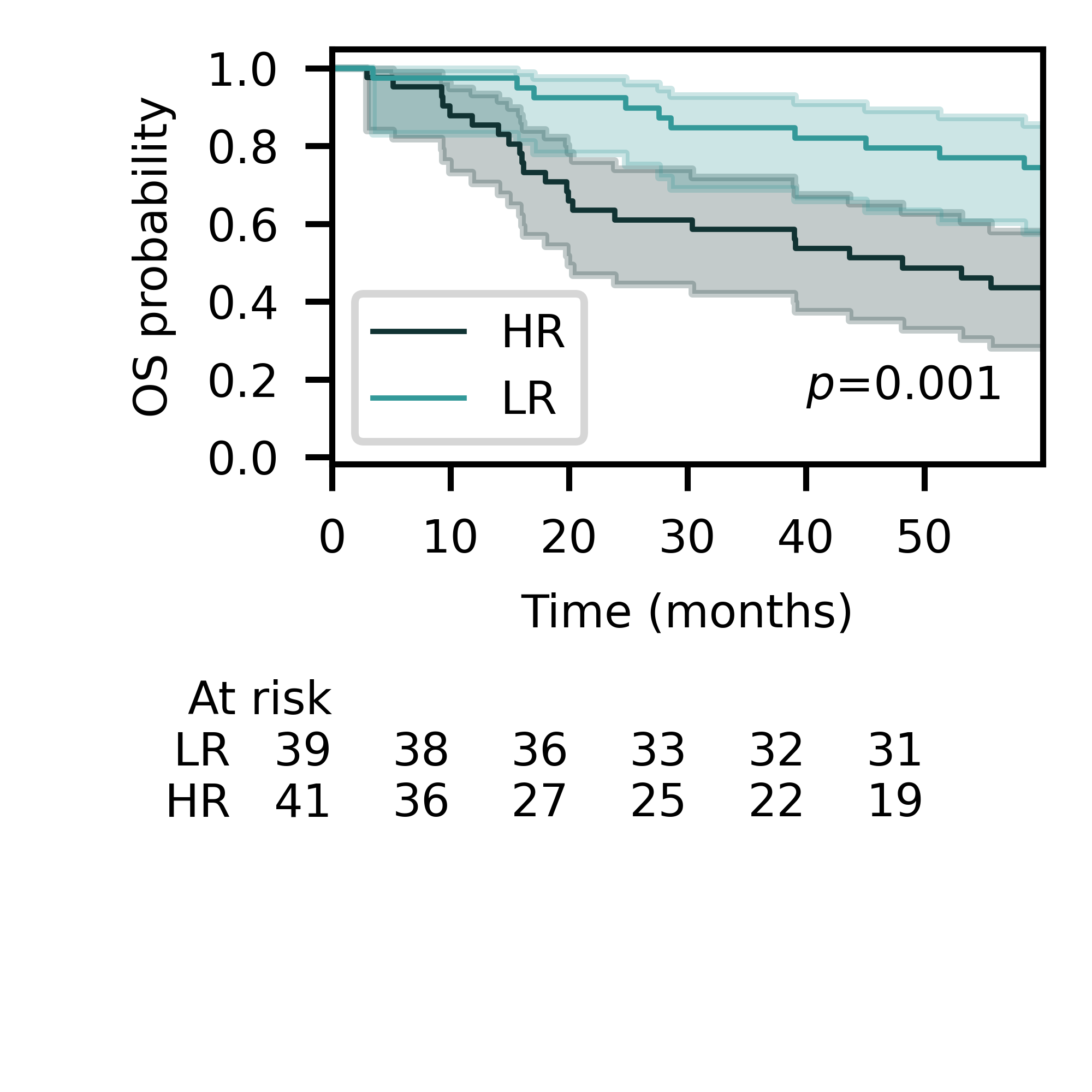}
    \includegraphics[width=0.32\textwidth]{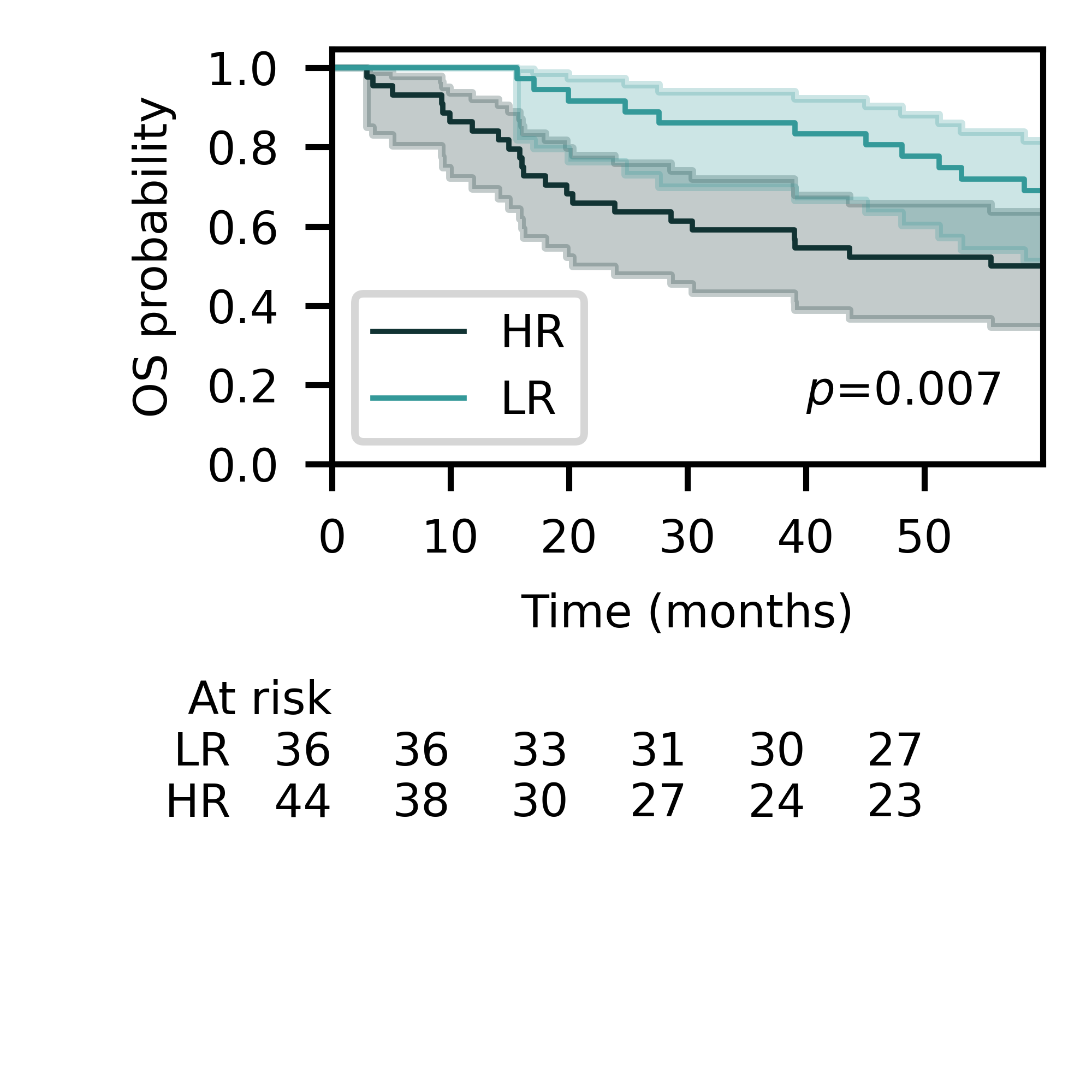}
    \includegraphics[width=0.32\textwidth]{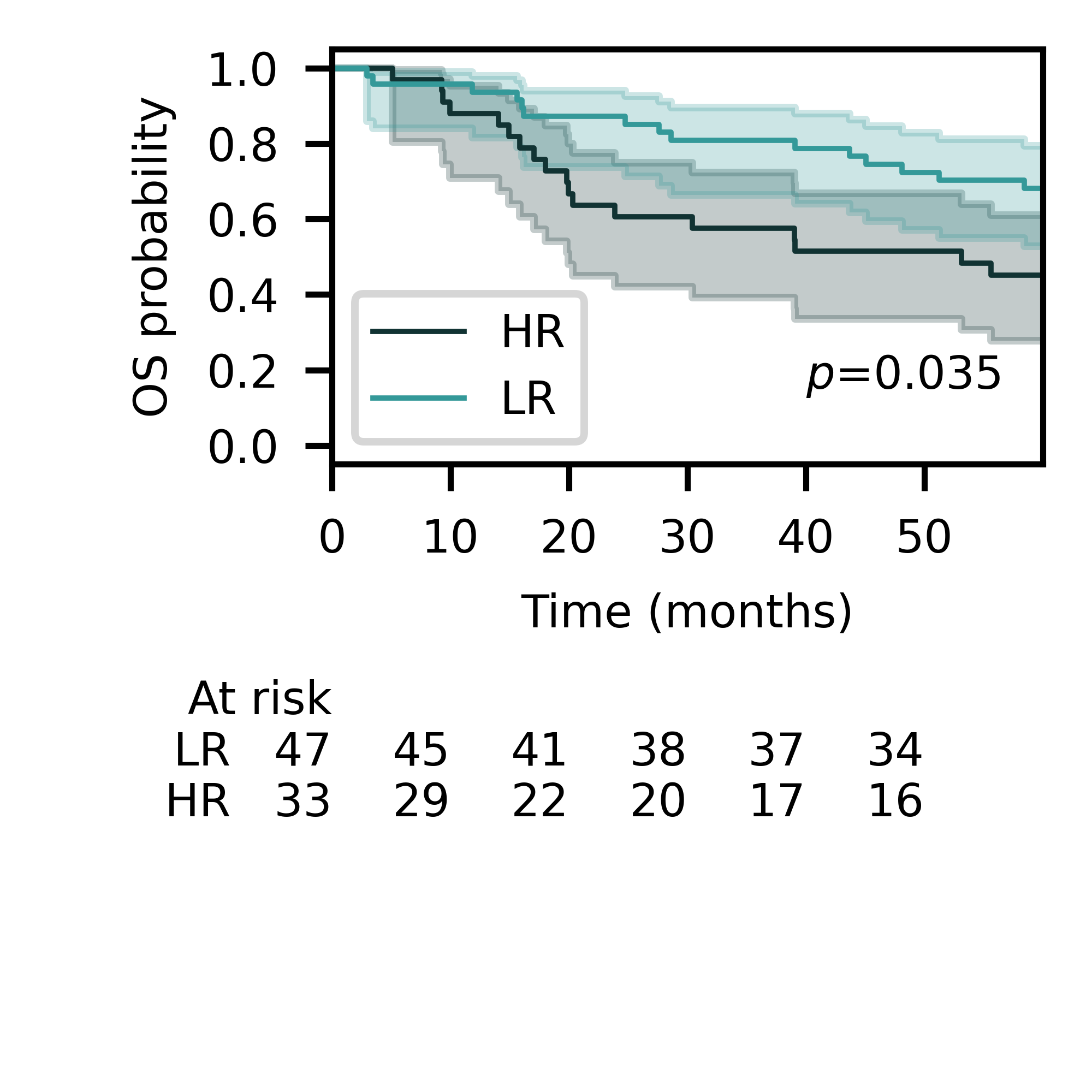}
    \end{minipage}
\vfill
    \begin{minipage}{\textwidth}
    \centering
    \includegraphics[width=0.32\textwidth]{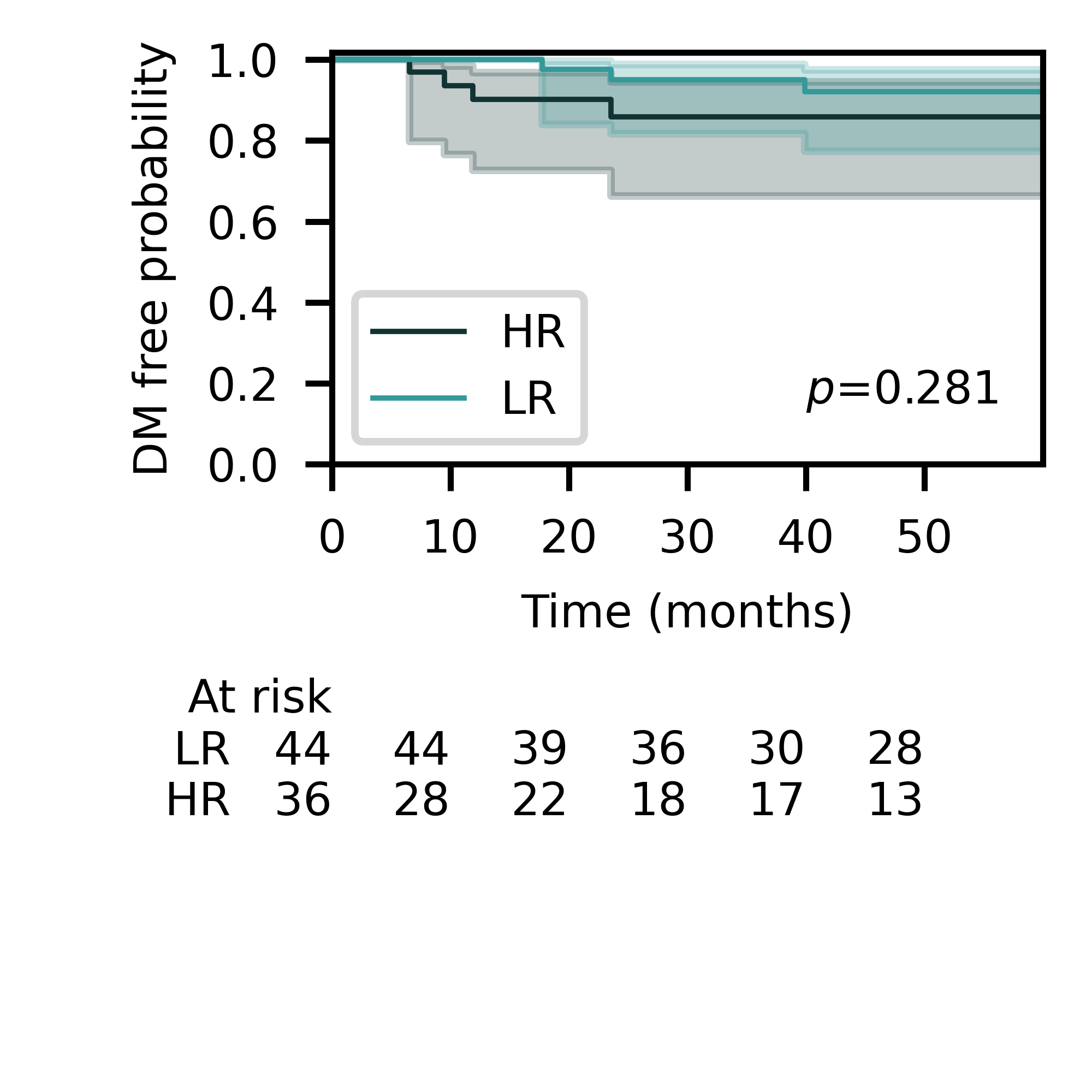}
    \includegraphics[width=0.32\textwidth]{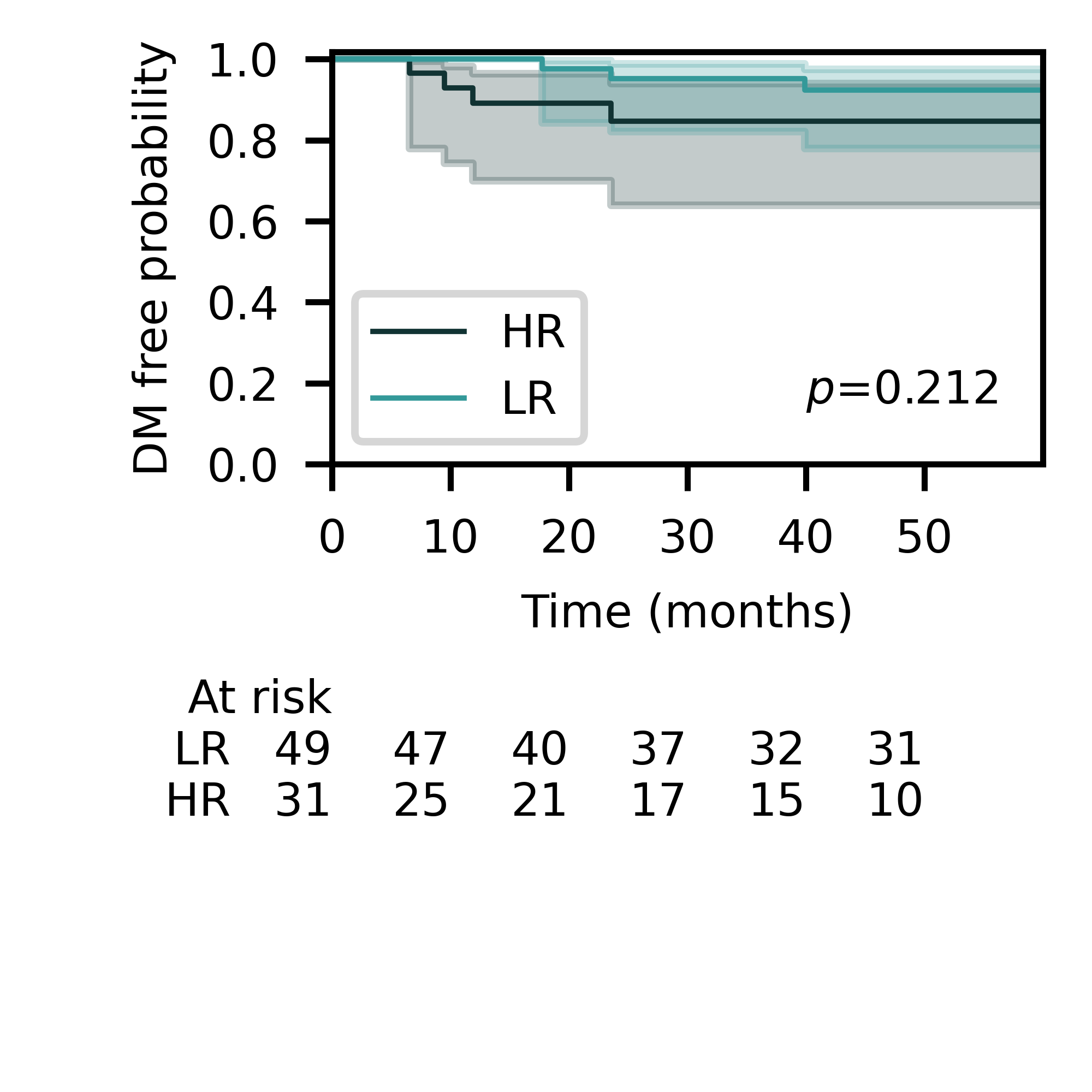}
    \includegraphics[width=0.32\textwidth]{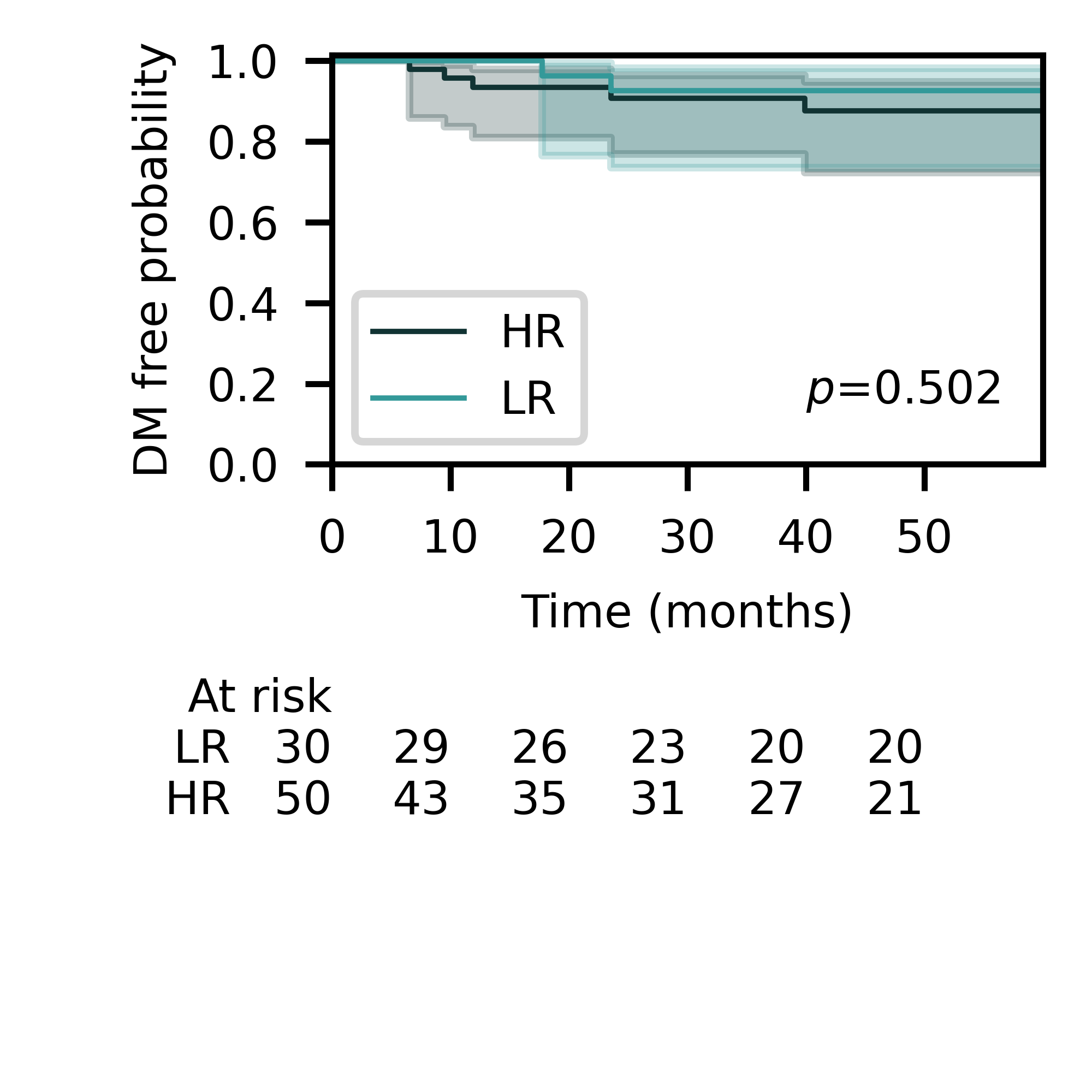}
    \end{minipage}
\caption{Results of time-to-event analyses for the test cohort. Top: OS, Bottom: DM; Left to right: Linear, LPNL and PHGN models.}
\label{km_tte}
\end{figure}

Figure \ref{km_tte} illustrates the results of all the models for time-to-event analyses. The linear model outperformed the GNN models by a margin of at least $0.03$ with a c-index of 0.68 for OS event and a c-index of 0.69 for DM event.  The c-indices for GNN models were: (\romannum{1}) LPNL: OS - $0.65$, DM - $0.64$; (\romannum{2}) PHGN: OS - $0.59$, DM - $0.65$. All models were able to achieve a statistically significant stratification of the test patient cohort for OS with $p<0.05$. However, none of the models were successful in achieving meaningful stratification for DM event. In the current study, radiomic features from lymph node segmentations were not considered for DM analysis in order to avoid further censoring of patient data. However, including this in future could help improve the performance of the models for DM analysis. Additional results are included in the supplementary material. 

\section{Discussion}
The PHGN model could not outperform the baseline linear model, while the LPNL model could not show a significant improvement over the linear model. This is in contrast to the results achieved by the baseline GNN models applied for brain MRI data \cite{cosmo2020latent,parisot2018disease}. Apart from the considered entity and modality, there are a few key differences between these existing methods and our methods. 

In the work of Parisot et al. \cite{parisot2018disease}, phenotypic information along with imaging features was used to construct the population graph. However, this phenotypic information was not included for the baseline linear model as per our knowledge. Since it is well established that the clinical features alone can have a huge impact on the outcome of the patient, the performance benefit observed in this work over the linear model might be due to the bias induced by phenotypic information. There was also a substantial impact on the performance of their GNN model when there was a small variation in the phenotypic information used. Therefore to carefully assess the benefit obtained from graph representations, clinical data was not included for all the models in this study. Another reason was to ensure uniformity across the different datasets used. 

For HPV status prediction task, features from only patients diagnosed with cancer was used in this study. The datasets used in baseline GNN studies \cite{cosmo2020latent,parisot2018disease} included both healthy control and diseased patients in the population graph. In addition, as suggested by Parisot et al. \cite{parisot2018disease} features extracted from a deep learning model such as Convolutional Neural Netwoks (CNNs) might be a better alternative to hand-crafted radiomics features in the construction of population graph based models. Furthermore, the study by Lang et al. \cite{lang2021deep} also suggests that the area outside the tumour volume, which is generally not captured by the radiomic features, might be beneficial for HPV status prediction. They draw this conclusion based on the performance benefit achieved by 3D CNN models over traditional radiomic models. It would be therefore interesting in future to include healthy controls in the patient cohort, and train CNN-GNN models in an end-to-end manner for outcome prediction.

One of the reasons for the low performance of PHGN model could be insufficient training data ($<500$ per fold) for learning a general rule and meaningful graph representations. The low performance of the model could also be due to the use of complex models for data that does not contain enough predictive power. That is, pre-treatment CT data for head and neck cancers have been speculated to contain limited predictive power \cite{starke20202d}. Therefore, the use of data from other modalities like MRI or histopathological imaging in combination with CT data; or the use of features from CT data over the entire course of treatment (delta radiomics) could be a better means to assess the performance of these complex models \cite{lambin2017radiomics,leger2019ct,lock2017residual}. 

In summary, we developed a novel HGNN model (PHGN) that used CT-based radiomic features, and compared it with GNN (LPNL) and baseline linear model for various outcome prediction tasks. The graph models could not definitively surpass baseline linear model in any of the investigated tasks. Based on the obtained results, a few avenues for future research have been identified, which could improve outcome prediction in cancer patients.   

\section{Acknowledgements}
The research was carried out within the Surgomics project funded by the German Federal Ministry of Health, University Hospital Heidelberg and Uniklinikum Carl Gustav Carus, Dresden. Partners from industry: Karl Storz GmbH and Phellow Seven GmbH.

\bibliographystyle{splncs04}
\bibliography{references}

\end{document}

\pagenumbering{arabic}
\title{\Large{Supplementary material: Graph data modelling for outcome prediction in oropharyngeal cancer patients}}
%
%\titlerunning{Graph data modelling for outcome prediction in OPC patients}

\author{Nithya Bhasker et al.}
%\institute{
%\email{abc@xyz.com}}
\date{}

\maketitle 

\begin{table}[!h]
\centering
\caption{Parameters used for the extraction of radiomic features}
\begin{footnotesize}
\begin{tabular}{|c|c|}
\hline
Parameters & Values \\
\hline
Interpolation method and voxel size & Linear; $1\times1\times1$mm$^{3}$ \\
Intensity threshold for ROI resegmentation & $-150$ to $180$ HU \\
Discretisation method & fixed bin number, $n=32$\\
Spatial filters for image transformation & wavelet (coiflet-1), Laplacian of Gaussian (LoG)\\
LoG filter width, $\sigma$ & 5, 4, 3, 2, 1, 0.5\\
\hline
\end{tabular}
\end{footnotesize}
\end{table}

\begin{table}[!h]
\centering
\caption{Optimised hyperparameters for different methods and tasks with the observed median value and range over $n=100$ trials. Feature size in the table refers to the size of the ranked feature subset.}
\begin{footnotesize}
\begin{tabular}{|>{\raggedright\arraybackslash}p{1cm}|>{\raggedright\arraybackslash}p{2.5cm}|>{\raggedright\arraybackslash}p{2.8cm}|>{\raggedright\arraybackslash}p{2.8cm}|>{\raggedright\arraybackslash}p{2.8cm}|>{\raggedright\arraybackslash}p{2.8cm}|}
\hline
Method & Hyperparameters & \multicolumn{4}{c|}{Observed median value and range} \\
\cline{3-6}
& & HPV & bin OS & OS & DM \\
\hline
\multirow{5}{1.3cm}{Linear models} & feature size & 197, [10, 329]& 167, [2, 328] & 193, [2, 334] & 168, [1, 336]\\
& tolerance for stopping criteria & 6e-4, [1.5e-5, 9.9e-4] & 4e-4, [2.5e-5, 9.7e-4] & 5.4e-4, [3e-5, 9.9e-4] &  6.6e-4, [1.3e-5, 9.9e-4]\\
& regularisation strength & 2.6, [3.5e-3, 4.95] & 2.94, [0.14, 4.95]& - & - \\
& iterations & 2556, [120, 4956] & 2167, [115, 4984]& 5.5e5, [1.5e3, 9.9e5] & 4.7e5, [1.6e4, 9.98e5]\\
& elastic net mixing parameter & 0.48, [0.15, 0.89]& 0.6, [0.1, 0.9]& 0.5, [0.1, 0.86] & 0.55, [0.13, 0.9]\\
\hline
\multirow{7}{1.3cm}{LPNL} & feature size & 177, [19, 335] & 185, [5, 333] & 156, [5, 332] & 161, [3, 334] \\
& size of MLP layers & 128, [5, 255]  & 152, [5, 253] & 144, [7, 255] & 139, [1, 251] \\
& size of GNN layers & 144, [16, 248] & 132, [6, 248] & 149, [8, 255] & 154, [3, 256] \\
& depth of hidden GNN layers & 39, [3, 63] & 32, [1, 64] & 32, [2, 64] & 40, [2, 62] \\
& dropout & 0.34, [6.8e-3, 0.77] & 0.41, [1.7e-3, 0.79] & 0.37, [4.4e-3, 0.8] & 0.45, [0.01, 0.79] \\
& learning rate (Adam) & 5.3e-4, [1.6e-5, 9.8e-4] & 7.1e-4, [1.9e-5, 9.8e-4] & 5.9e-4, [1.5e-5, 9.9e-4] & 4.2e-4, [3.5e-5, 9.9e-4] \\
& decay (Adam) & 3.9e-4, [2.8e-5, 9.8e-4] & 5e-4, [2.8e-5, 9.9e-4] & 5.4e-4, [5.7e-5, 9.9e-4] & 5.2e-4, [1.1e-5, 9.9e-4] \\
\hline
\multirow{9}{1.3cm}{PHGN} & feature size & 219, [6, 330]  & 204, [16, 336]  & 160, [1, 325] & 177, [5, 331] \\
& $k$ for k-nn & 3, [3, 5] & 4, [3, 5] & 4, [3, 5] & 4, [3, 5] \\
& size of dense layer (input) & 118, [6, 253] & 135, [7, 256] & 96, [10, 250] & 136, [3, 249] \\
& size of dense layer (output) & 140, [13, 256] & 167, [10, 246] & 158, [1, 250] & 129, [1, 247] \\
& size of HGNN layers & 143, [3, 244] & 144, [6, 253] & 127, [8, 249] &  115, [7, 256] \\
& depth of hidden HGNN layers & 38, [2, 64] & 27, [1, 64] & 25, [1, 64] & 31, [2, 62] \\
& dropout & 0.32, [8.1e-3, 0.78] & 0.38, [1.8e-3, 0.8] & 0.36, [9.6e-5, 0.8] & 0.47, [0.01, 0.8] \\
& learning rate (Adam) & 6e-4, [3.3e-5, 9.9e-4] & 5.9e-4, [3.6e-5, 9.9e-4] & 5.3e-4, [1.2e-5, 9.9e-4] & 4.7e-4, [1.5e-5, 9.9e-4] \\
& decay (Adam) & 4e-4, [1.2e-5, 9.9e-4] & 5.3e-4, [3e-5, 9.8e-4] & 4.7e-4, [1.5e-5, 9.9e-4] & 6.5e-4, [1.7e-5, 1e-3] \\
\hline
\end{tabular}
\end{footnotesize}
\end{table}

\begin{figure}[!h]
    \centering
    \includegraphics[width=0.32\textwidth]{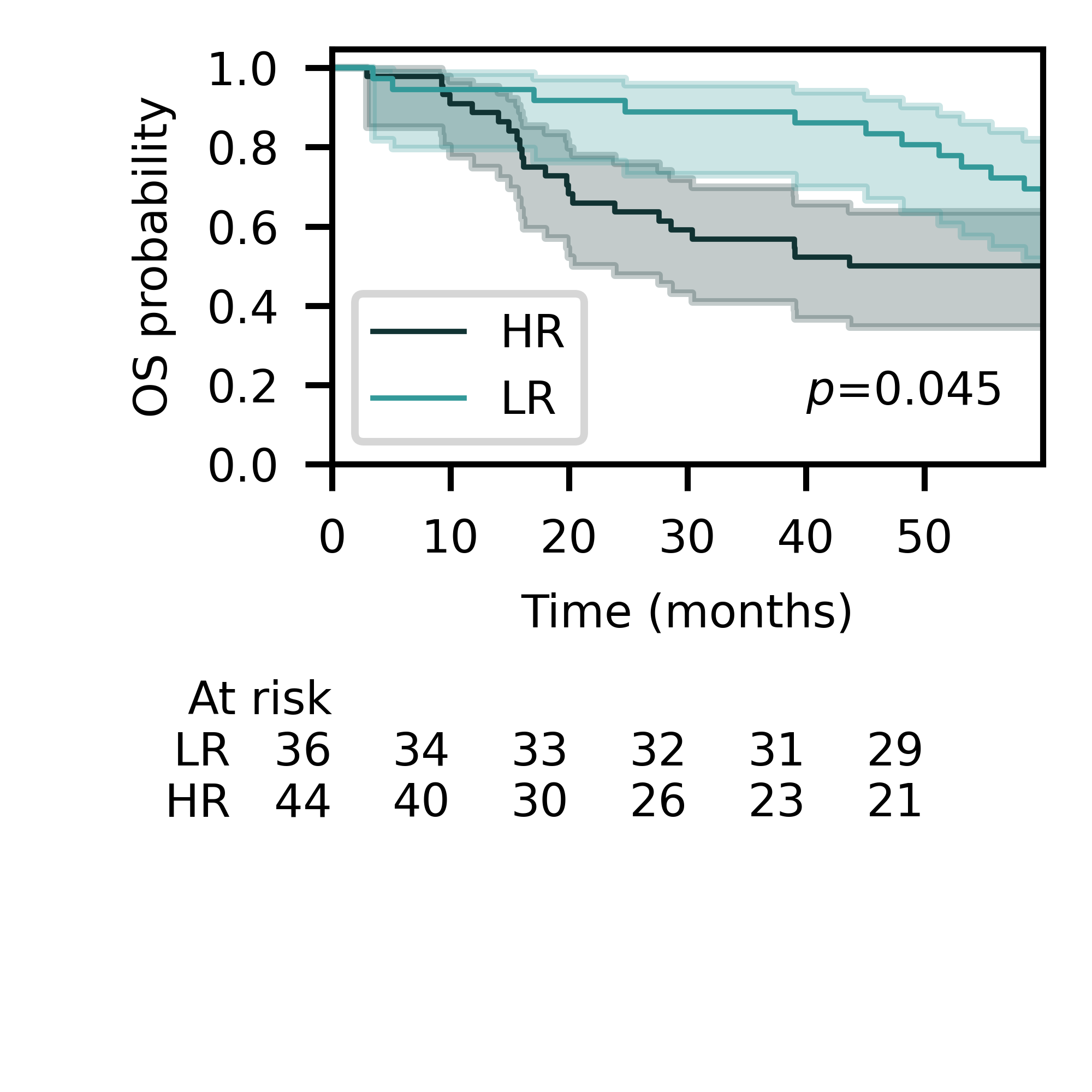}
    \includegraphics[width=0.32\textwidth]{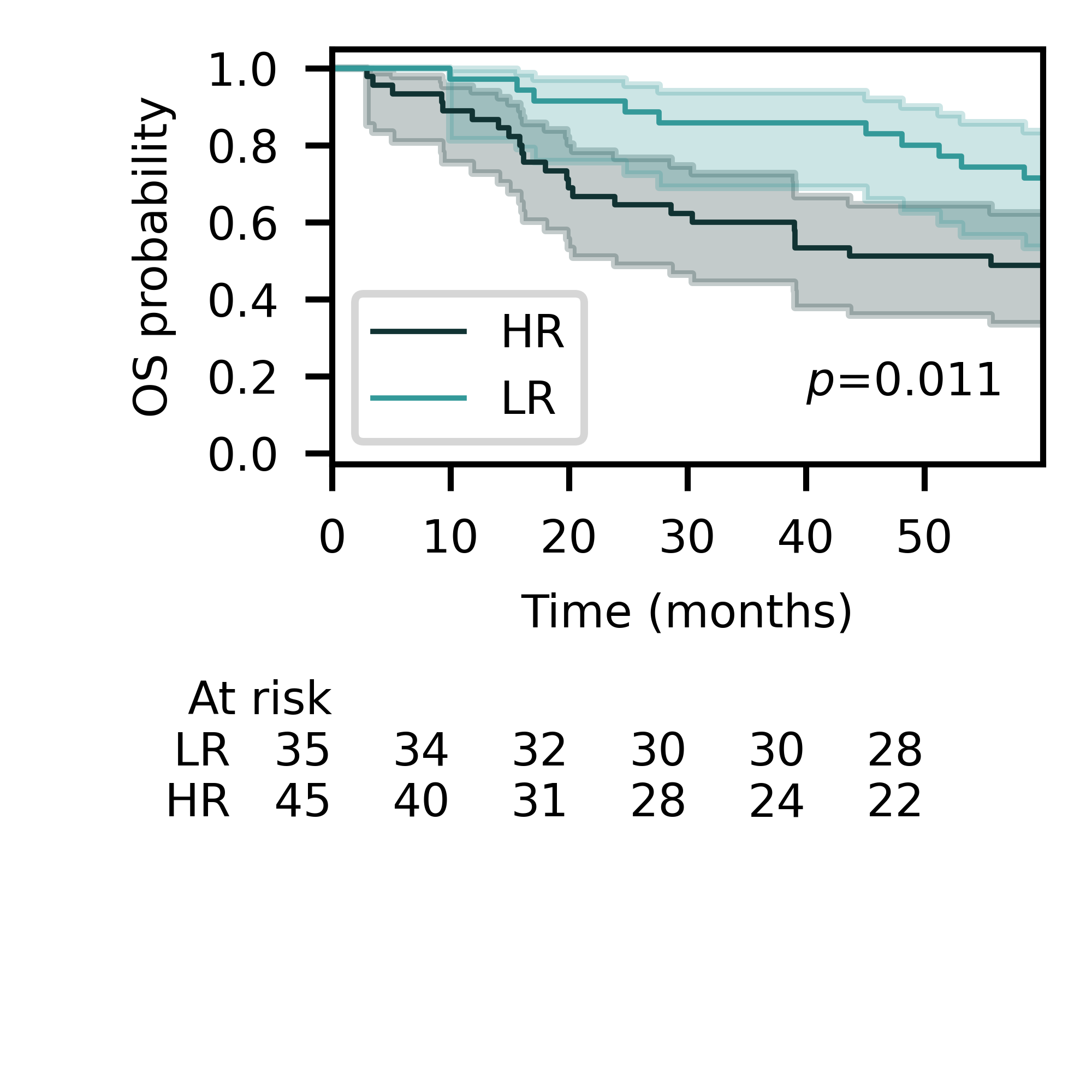}
    \includegraphics[width=0.32\textwidth]{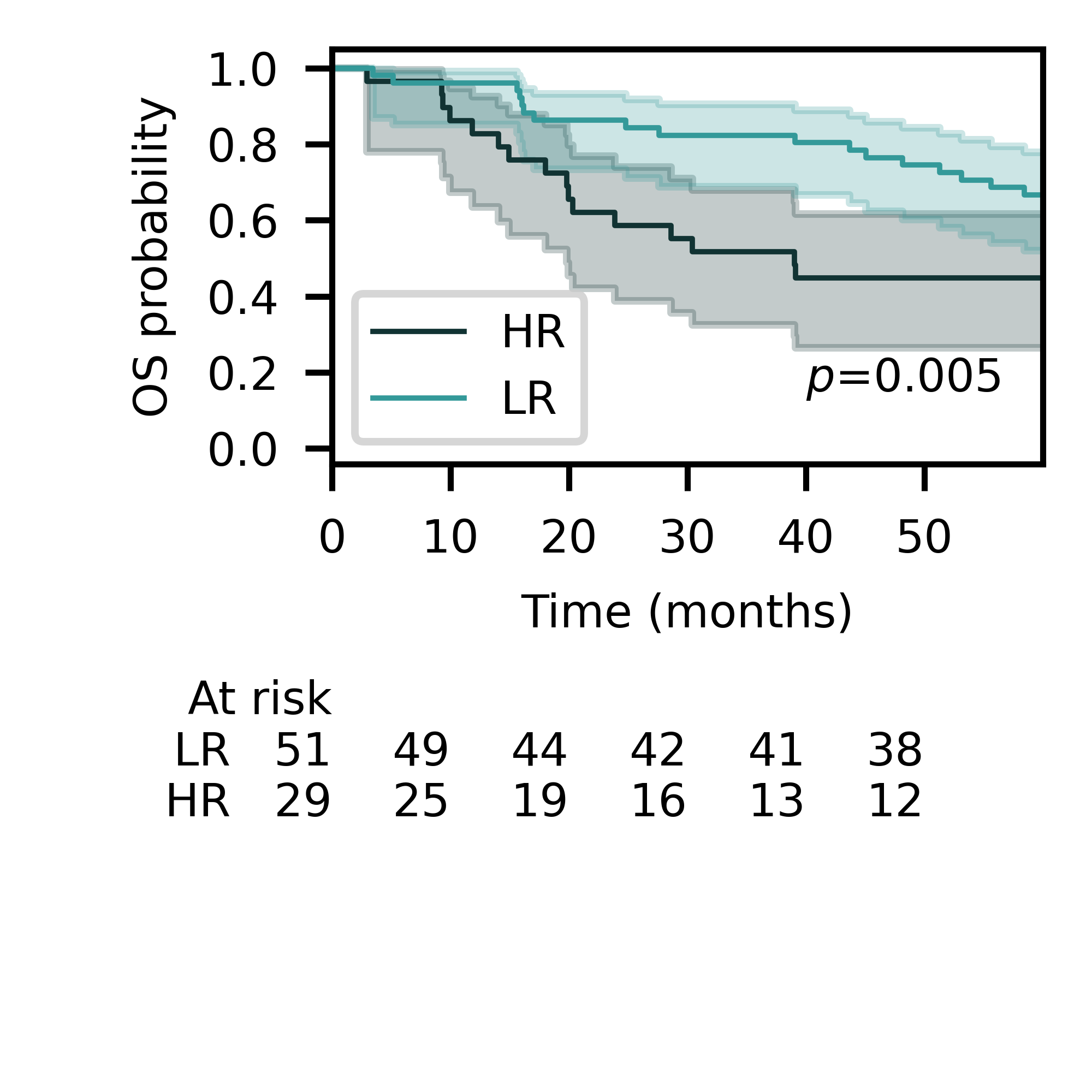}
\caption{Kaplan-Meier plots for binary OS prediction on the test cohort. HR - high risk group; LR- Low risk group. Left to right: Linear, LPNL and PHGN models.}
\label{km_bin}
\end{figure}

\begin{figure}[!h]
    \centering
    \includegraphics[width=0.4\textwidth]{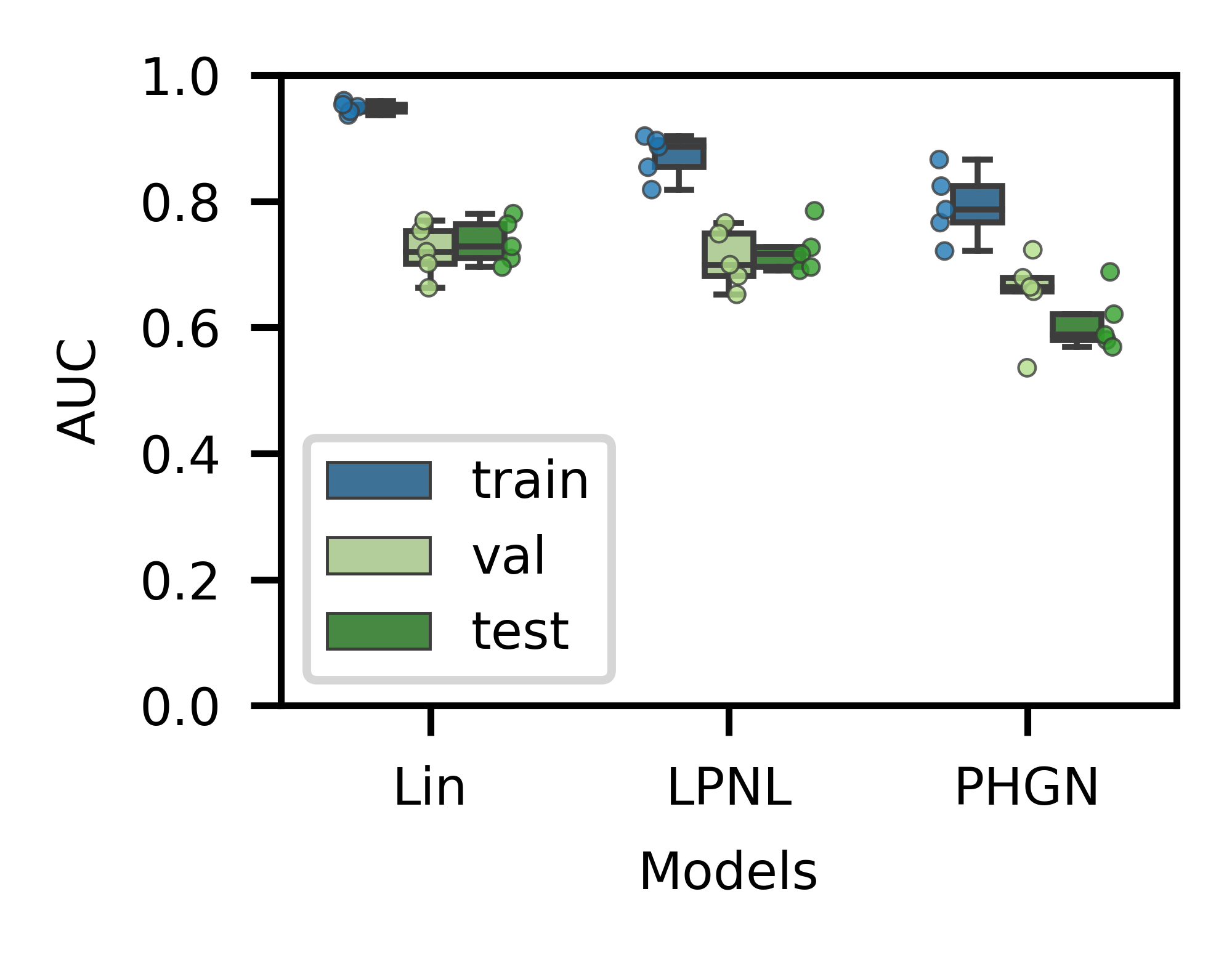}    
    \includegraphics[width=0.4\textwidth]{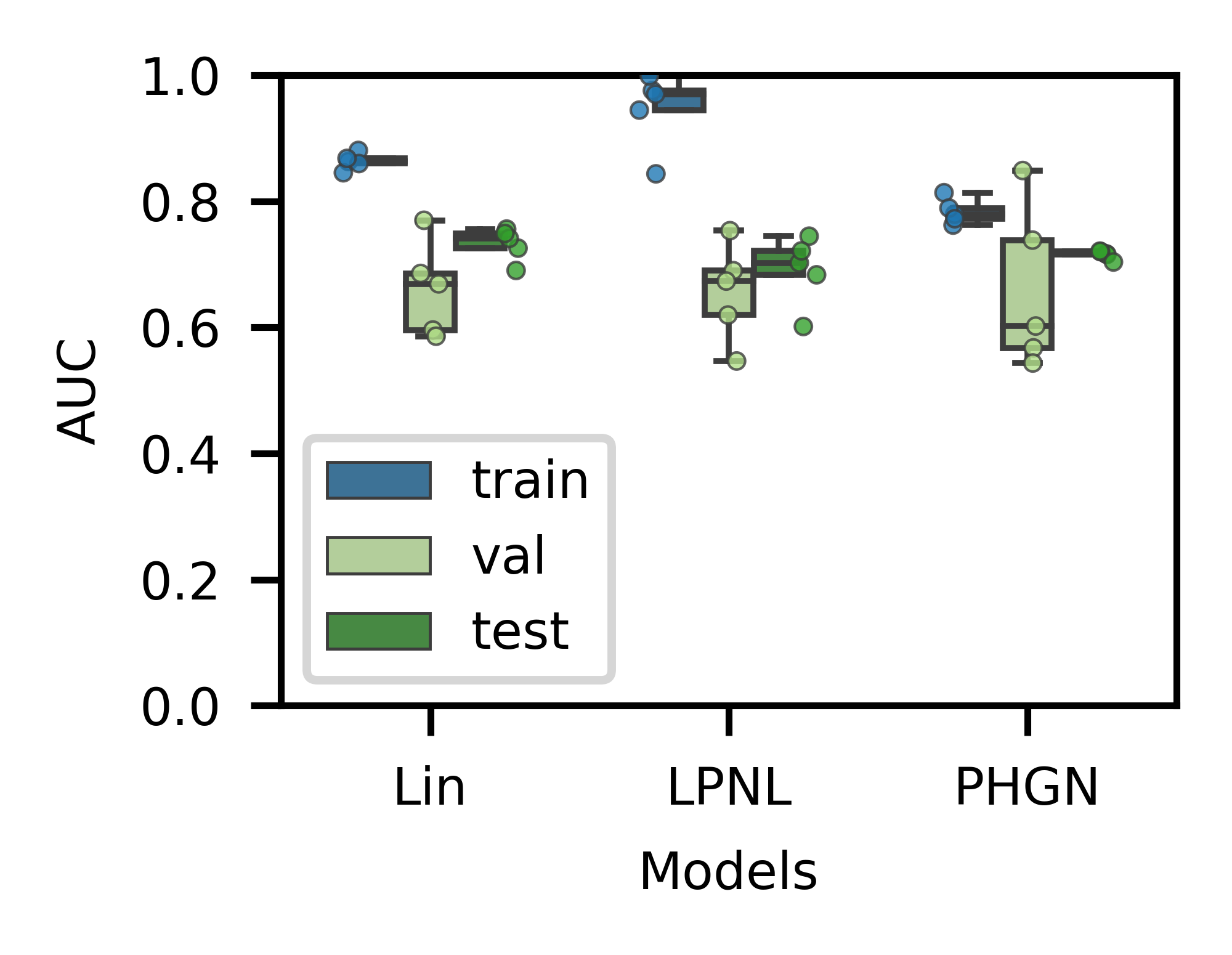}
    \includegraphics[width=0.4\textwidth]{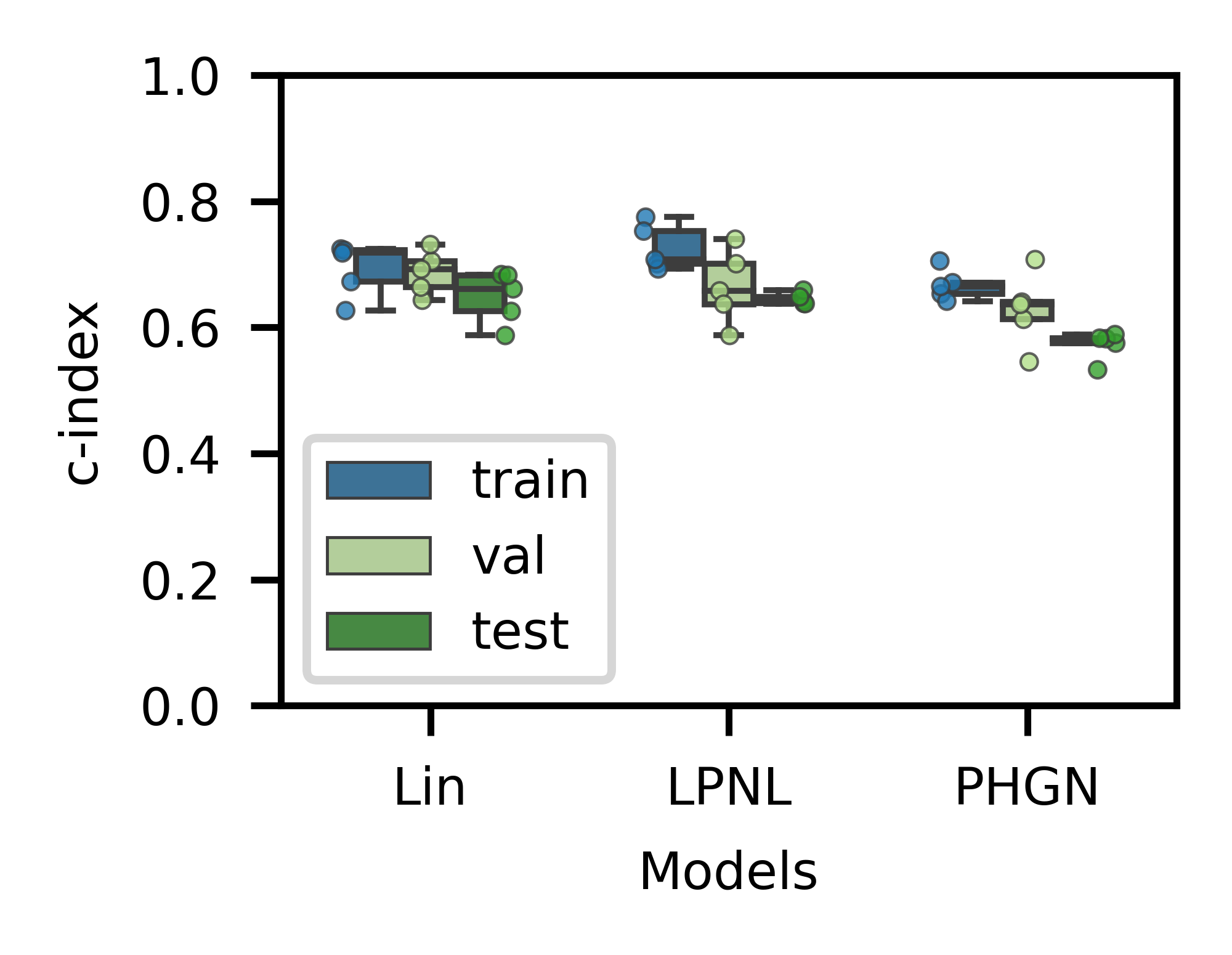}
    \includegraphics[width=0.4\textwidth]{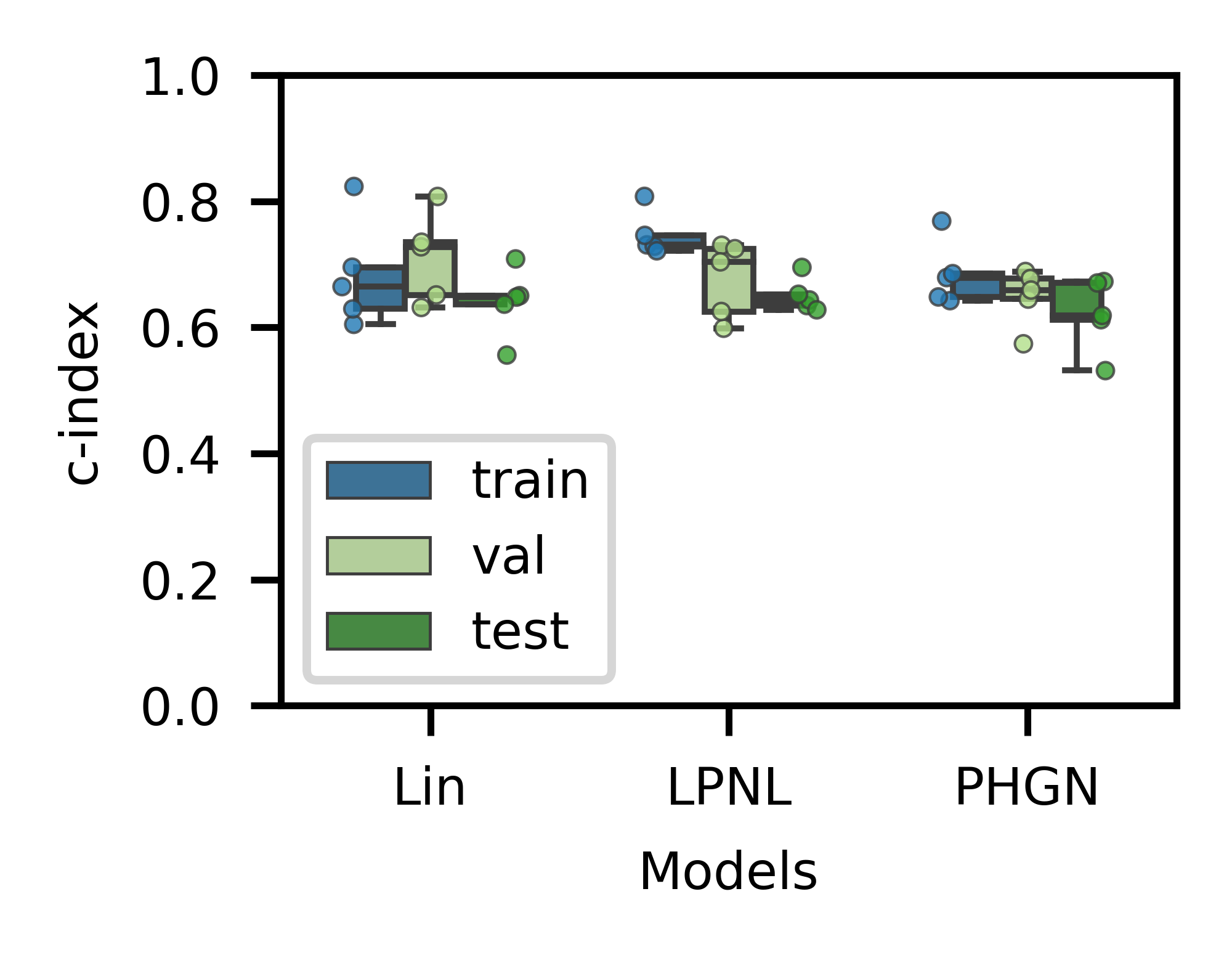}
\caption{Performance metrics across the cross validation folds for different tasks. Top: Left - HPV, Right - binary OS; Bottom: Left - OS, Right - DM}
\label{cv}
\end{figure}